\documentclass[letterpaper, 10 pt, conference]{ieeeconf}  

\IEEEoverridecommandlockouts                              

\usepackage{balance} 
\usepackage[utf8]{inputenc}
\usepackage{amsfonts}
\usepackage{amsmath}
\usepackage{xcolor}
\usepackage{graphicx}
\usepackage{multirow}
\usepackage{units}
\usepackage{tabularx}
\usepackage{paralist}
\usepackage{booktabs} 
\usepackage{hyperref}

\DeclareMathOperator*{\argmax}{arg\,max}

\title{\LARGE \bf
A Study of Reinforcement Learning Algorithms\\
for Aggregates of Minimalistic Robots}

\author{Joshua Bloom$^{1}$, Apratim Mukherjee$^{1}$, and Carlo Pinciroli$^{1}$
\thanks{$^{1}$Department of Robotics Engineering,
        Worcester Polytechnic Institute, Worcester, MA 01609, USA.
        E-mail: {\tt\small \{jdbloom, amukherjee3, cpinciroli\}@wpi.edu}.}%
}

\begin{document}

\maketitle
\thispagestyle{empty}
\pagestyle{empty}

\begin{abstract}
The aim of this paper is to study how to apply deep reinforcement learning for the control of aggregates of minimalistic robots. We define \textit{aggregates} as groups of robots with a physical connection that compels them to form a specified shape. In our case, the robots are pre-attached to an object that must be collectively transported to a known location. Minimalism, in our setting, stems from the barebone capabilities we assume: The robots can sense the target location and the immediate obstacles, but lack the means to communicate explicitly through, e.g., message-passing. In our setting, communication is implicit, i.e., mediated by aggregated push-and-pull on the object exerted by each robot. We analyze the ability to reach coordinated behavior of four well-known algorithms for deep reinforcement learning (DQN, DDQN, DDPG, and TD3).
Our experiments include robot failures and different types of environmental obstacles. We compare the performance of the best control strategies found, highlighting strengths and weaknesses of each of the considered training algorithms.
\end{abstract}



\section{INTRODUCTION}

An interesting niche of swarm systems is that of \emph{aggregates}, i.e., teams of robots connected to each other by mechanical links. Modular robots \cite{sunSalamanderbotSoftrigidComposite2020}, self-assemblying robots \cite{mondadaSwarmBotNewDistributed2004}, and certain approaches to collective transport \cite{tuci_cooperative_2018} fall into this category. Controlling robotic aggregates is, in general, not trivial, due to the need for coordinated pushing and pulling across the individual robots.



In this paper, we show that multi-agent reinforcement learning (MARL) is a promising approach to tackle the control of robotic aggregates. A common problem in MARL is non-stationarity due to the independent actions of each robot. This is often solved through careful message-passing and advanced training schemas \cite{amato_modeling_2019,foerster_learning_2016,wu_spatial_2021}.

However, in this paper, we show that effective coordination can be obtained with well-known RL algorithms even when explicit communication or sharing of action/observations is not possible. To support this insight, we consider a collective transport scenario in which a swarm of minimalistic robots must move an object to a predefined location. We assume the robots able to navigate and acquire basic obstacle proximity readings, but unable to communicate directly through, e.g., message-passing. Crucially, we also assume that the robots are physically constrained to be connected to an object that must be carried to a target location. The robots can also sense the direction to this location.

This minimalistic collective transport task is a compelling testbed for decentralized deep reinforcement learning. Because of their inability to communicate and of the physical constraint that forces them to coordinate transport, the robots must learn to push and pull, as well as to model other robots' actions. This is a form of implicit communication that is based on the compounded actions of all robots, which effectively circumvents the problem of non-stationarity. In essence, for each robot, training boils down to modeling the behavior of the object as a disturbance with respect to its individual actuation.

To demonstrate the viability of our insight, we compare the performance of four well-known algorithms for reinforcement learning: DQN \cite{mnih_human-level_2015}, DDQN \cite{van_hasselt_deep_2016}, DDPG \cite{lillicrap_continuous_2015}, and TD3 \cite{fujimoto_addressing_2018}. In all methods, each robot is equipped with a single deep neural network, without explicit modules to model other robots.

Our results indicate that, despite the barebone minimalism of our robots, using the object as a medium for coordination produces solutions that display scalability, resilience to failures, and obstacle avoidance. We analyze the behavior and performance of the different behaviors produced by the training algorithms through a thorough campaign of experiments.


\section{Background and Related Work}
\subsection{Reinforcement Learning}
Reinforcement Learning (RL) problems are formalized as Markov Decision Process (MDP) represented as the tuple $\langle S, A, R, T \rangle$, where $S$ is a set of states, $A$ is the set of actions, $R(s, a, s'): S \times A \times S \mapsto \mathbb{R}$ is the reward function, and $T(s' | s, a)$ is a probabilistic function mapping states and actions to a new state. MDPs are formalized using the Bellman Equation:
\begin{multline}\label{Bellman}
    Q(s,a) = \sum_{s'} \, T(s' | s,a) \, [ \, R(s,a,s') +\\
    + \gamma \, \argmax_{a'} Q(s',a') \, ]
\end{multline}
where $Q(s, a)$ is a function reflecting expected future rewards and $\gamma$ is a discount factor.

Applying RL to robotics introduces complexities as the robots perceive the environment through sensors and, in most cases, only receive a partial observation of the true state of the environment. Therefore, the MDP is extended to a Partially Observable MDP (POMDP) represented by the tuple $\langle S, A, R, T, \Omega, O \rangle$, where $o\in \Omega(s)$ is a partial observation of the full state $s \in S$ according to a probabilistic function $O(o | s)$.

In real-world problems the state-action space can become quite large and finding the true values of $Q$ difficult. Recently, neural networks (NN) have emerged as a viable method to approximate the value function $Q(s,a)$. Two of the most common approaches are value-based and policy-based RL. 

Mnih \emph{et al.} \cite{mnih_human-level_2015} introduced Deep Q-Network (DQN) which uses a deep neural network to learn, via \eqref{Bellman}, the mapping between states and the expected future rewards of actions. This method proved highly successful, showing comparable results in several Atari games to human-level performance. 

DQN, however, suffers from overestimation bias of the $Q$-values resulting in unstable convergence in the partially observable setting. To address this bias, van Hasselt \emph{et al.} \cite{van_hasselt_deep_2016} developed Double Deep Q-Network (DDQN) which uses a decoupled copy of the current $Q$-network (called $Q'$) to estimate the best action to take. This decoupled copy is updated frequently to reflect the current weights of the $Q$-network. While value-based methods are powerful in their ability to learn highly complex mappings between states and actions, they might suffer from long convergence times.

Policy-based methods, on the other hand, work directly on the mapping from states $s\in S$ to actions $a \in A$, called the \emph{policy} $a = \mu(s)$. Lillicrap \emph{et al.} \cite{lillicrap_continuous_2015} applied NNs to Deterministic Policy Gradient (DDPG), which uses an actor to learn a policy $\mu$ and a critic to evaluate the policy of the actor through a value function $Q(s, a)$.

Similar to DQN, DDPG suffers from overestimation bias in its value function, $Q$, leading to inflated estimates of expected future reward. Fujimoto \emph{et al.} \cite{fujimoto_addressing_2018} employ a two-critic evaluation into their algorithm Twin Delayed DDPG (TD3) to minimize this overestimation bias. When evaluating the policy $\mu$, two evaluation values are calculated by separate critics and the minimum between them is chosen.

\subsection{Multi-Agent Reinforcement Learning}

Multi-Agent Reinforcement Learning (MARL) can be formalized as a Decentralized POMDP (Dec-POMDP) $\langle I, S, \{A_i\}, R, T, \{\Omega_i\}, O \rangle$ where $I$ is the set of agents, $A = \times_i A_i$ is the set of joint actions, and $\Omega = \times_i \Omega_i$ is the set of joint observations. Dec-POMDPs typically assume communication, synchronization, or observability over all agents within the environment to avoid non-stationarity \cite{foerster_learning_2016}, \cite{qie_joint_2019}, \cite{xiao_learning_2020}. Non-stationarity stems from the fact that the agents consider their neighbors as part of the environment. As a consequence, the behavior of the individual agents becomes interdependent and learning convergence is not guaranteed. A typical solution to non-stationarity is to require the agents to share information. Amato \emph{et al.} \cite{amato_modeling_2019} achieved this via consensus on a joint action space. Foerster \emph{et al.} \cite{foerster_learning_2016} proposed to communicate differentiable units as feedback for training. Wu \emph{et al.} \cite{wu_spatial_2021} built a belief state via spatial intention maps by directly observing physical locations of nearby agents over time.

Aggregates, however, provide a physical connection that allows us to forego any explicit information sharing, be it communication or direct observation. This form of indirect communication is an interesting unexplored coordination avenue in reinforcement learning.

\subsection{MARL for Collective Transport}

The idea of applying reinforcement learning to aggregate and modular robots \cite{liuLearningLocomoteArtificial2020}, and in particular to collective transport, has already been explored in the recent literature. However, to the best of our knowledge, none of the existing works share the same minimalistic assumptions that we study in this paper.

Zhang \emph{et al.} \cite{zhang_decentralized_2020} employs DDQN in a 2-robot transportation task in which the robots are constrained to be attached to an oversized rod. The robots must pick an action from a set of 4 predefined options  (forward-left, backwards-left, etc.). Under the assumption of perfect knowledge of pose and velocity of other agents, this approach produces control strategies able to transport the rod successfully through a door obstacle.

Other approaches to collective transport do not constrain the robots to the object, and instead rely on pushing. While they do not directly relate to our aggregate setting, they provide interesting insight on how to use the box to progress with transport. Wang \emph{et al.} \cite{wang_multi-robot_2006} and Rahimi \emph{et al.} \cite{fujita_comparison_2019} applied Q-learning to a box-pushing task. The robots are assumed able to push and must learn where to push the box by picking from a discrete set of predefined pushing locations. Eoh \emph{et al.} \cite{eoh_cooperative_2021} used DQN with a multi-agent box-pushing task in which they compare independent learners and a policy-reuse scheme. While the simplicity of box-pushing lends itself well to an RL formulation, the fact that the robots policy is changing while training makes the problem non-stationary.

\section{Methodology}

\subsection{Collective Transport}
We assume a cylindrical object to be transported and that the robots are uniformly distributed around the object. Transporting a cylindrical object with uniformly distributed robots allows us to make the assumption that we have symmetrical forces acting on the object. However, we allow robots to fail during testing and training, which introduces asymmetry. 

\begin{figure}
  \centering
  \includegraphics[width=\linewidth]{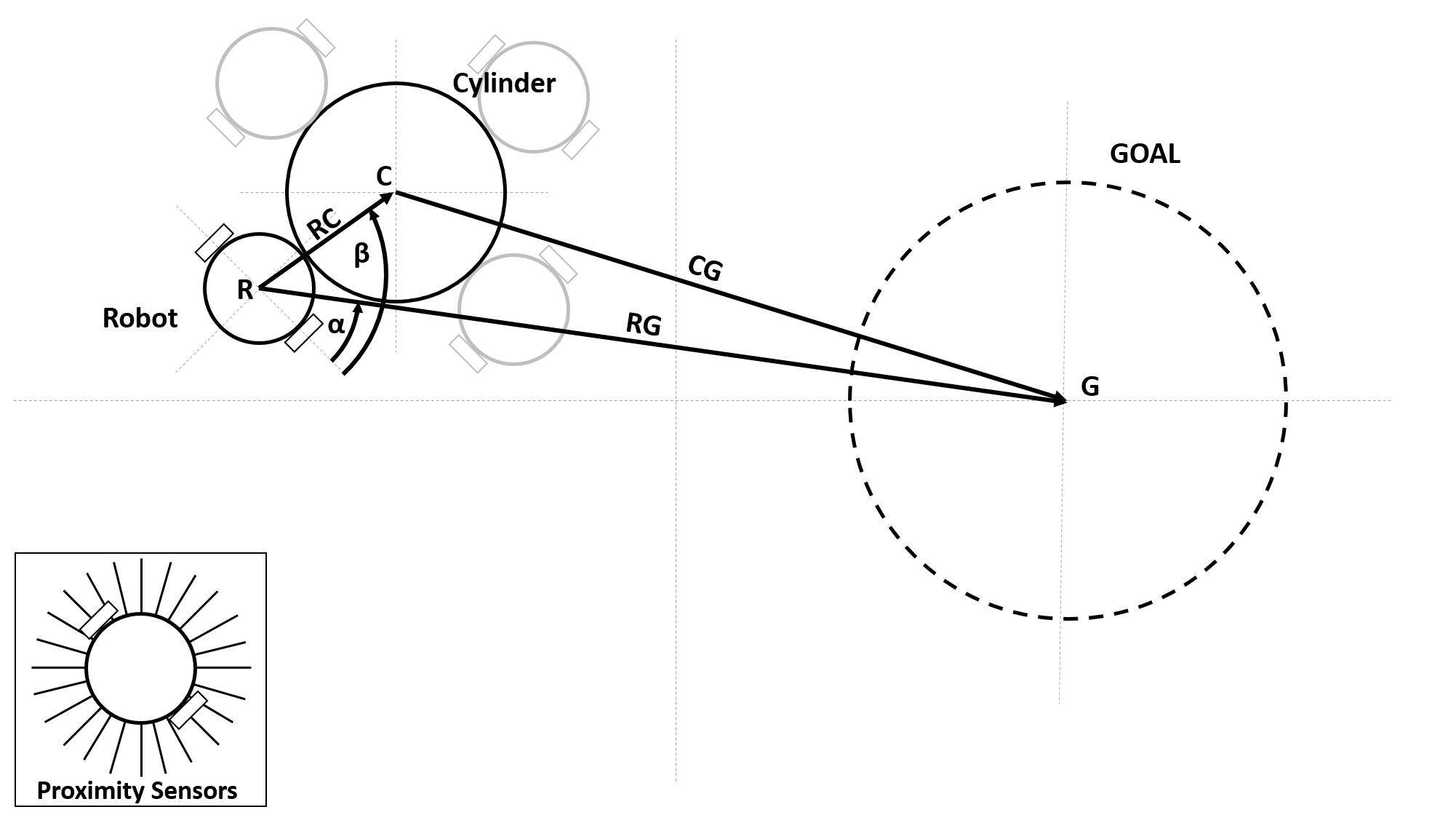}
  \caption{The robots have access to several spatial observations including the distances and angles to the cylinder and the goal from themselves. They also have access to the distance from the cylinder to the goal and an array of 24 proximity sensor readings that are uniformly distributed around the robot.}
  \label{fig:Robot-observation}
\end{figure}

Each robot is able to observe the tuple $\langle \overrightarrow{RG}, \overrightarrow{RC}, \overrightarrow{CG}, \textbf{p}\rangle$ according to its local frame of reference as shown in Figure \ref{fig:Robot-observation}, $\overrightarrow{RG}$ is the vector from the robot to the goal, $\overrightarrow{RC}$ is the vector from the robot to the cylinder, $\overrightarrow{CG}$ is the vector from the cylinder to the goal, and $\textbf{p}$ is an array of 24 proximity sensor readings. The sensors are distributed uniformly around the robot, and have a sensing range of $\unit[2]{m}$. The values returned from the sensors are normalized between 0 and 1, where 1 indicates maximum proximity. The values fade to zero exponentially with the distance.

The robots act independently and they have no means to explicitly communicate information between them. However, they are all rigidly connected to the object they are transporting. From the viewpoint of a single agent, other robots' actions upon the object can be viewed as a discrepancy from its own action and the resultant state. We can then model the dynamics on the object as a single actor with error. All agents involved strive to minimize this error because they share the same goal and reward function. 

\subsection{Robot Control}
We used the foot-bot \cite{bonani_marxbot_2010}, a differential-drive robot with an independent non-actuated turret attached to an actuated gripper. The turret allows the base to rotate independently from the gripper. The gripper is not included in the control schema as it is only actuated upon initialization or failure and is not actively controlled by the robot. 

The robots are controllable through wheel velocities $v\in (-10, 10)$ \unit{cm/s}. The agent chooses a $\Delta v \in \{-0.1, 0, 0.1\}$ \unit{cm/s} in the case of value-based methods and a $\Delta v \in \left[-0.1, 0.1\right]$ \unit{cm/s} for policy-based methods. Note the distinction between a discretized action space in the former and a continuous action space in the latter. In the case of value-based methods, we must provide all possible combinations between the two wheels as output, represented as $|\Delta v|^2$. We chose $|\Delta v|=3$ for control simplicity. 

\subsection{Obstacles}
\begin{figure}
  \centering
  \includegraphics[width=\linewidth]{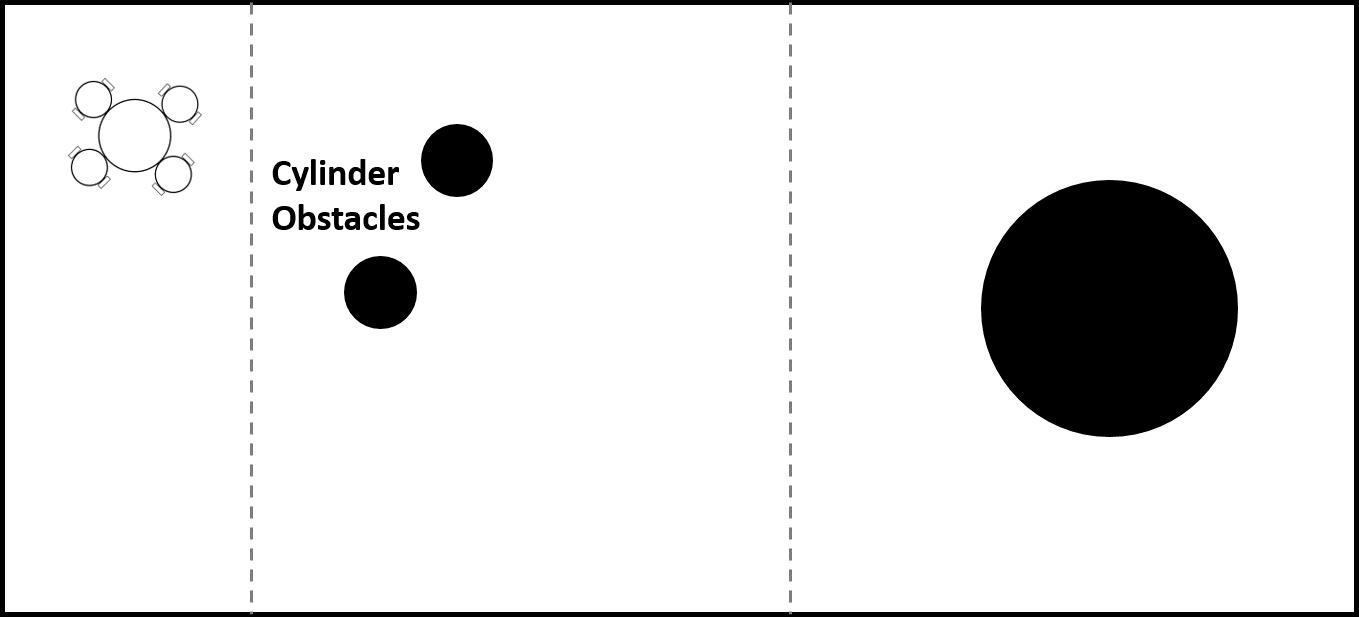}
  
  \vspace{0.1cm}
  
  \includegraphics[width=\linewidth]{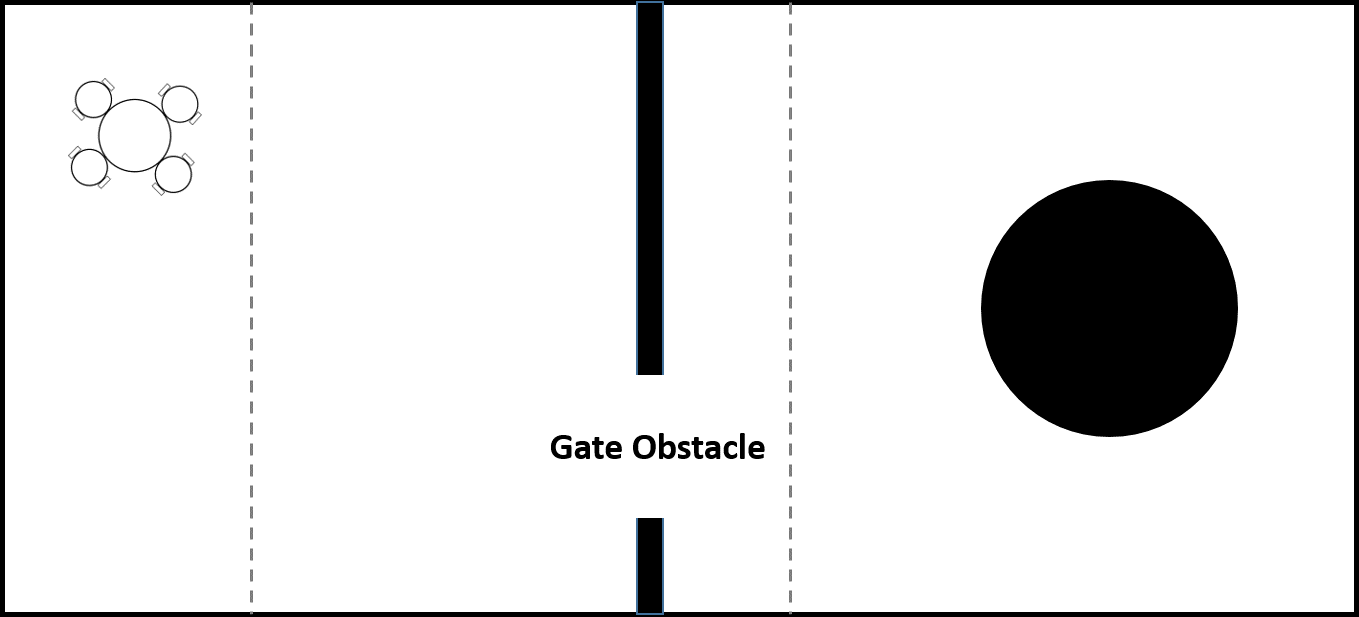}
  \caption{The top plot shows an environment with two cylinder obstacles. The bottom plot shows an environment with the gate obstacle. Robots and the object to transport are generated in the region on the left of the obstacles, the goal is generated in the region on the right.}
  \label{fig:Obstacles}
\end{figure}
Obstacles, shown in Figure \ref{fig:Obstacles}, are defined as stationary non-movable rigid objects. We generate all obstacles according to a uniform distribution. The cylinder obstacles are generated by selecting a $(x, y)$ position and the gate is generated by first selecting $x$ and then selecting the opening position $y$.

\subsection{Training}
We implement Centralized Training for Decentralized Execution (CTDE) \cite{lowe_multi-agent_2020} where experiences are generated from multiple agents and used to train a single centralized model offline. When execution time comes, this model is directly copied to all agents and operates solely through local experiences. Because this model was trained using local observations, it is agnostic to the number of agents it is applied to in execution. 

We use curriculum learning \cite{bengio_curriculum_2009} to enable our robots to learn how to navigate through the gate obstacle. We start with the gate opening equal to the width of the environment (effectively no obstacle). Then, we progressively shorten the width of the gate opening, until we reach our desired minimum gate opening. The latter is set to four times the diameter of the cylinder being transported to give room for the object and the robots connected to it.

We tested several different reward structures, in pursuit of a single structure that allowed us to learn environments with failures and obstacles. We settled on one that rewards moving in the direction of the goal while penalizing proximity and time taken to complete the task:
\begin{equation}
    R_{i, t} = -2 + \frac{\overrightarrow{CG}\cdot \bigg(C(x_t, y_t) - C(x_{t-1}, y_{t-1})\bigg)}{|\overrightarrow{CG}|\Bigg|\bigg(C(x_t, y_t) - C(x_{t-1}, y_{t-1})\bigg)\Bigg|} - \frac{\sum\textbf{p}}{|\textbf{p}|}
\end{equation}
where $\textbf{p}$ is the array of proximity values.

The combination of the direction of movement reward and the proximity reward as a general reward structure allows the algorithm to successfully scale, be resilient to failure, and clear certain obstacles, as discussed in Section \ref{Exp}. 

\subsection{Neural Network Architecture}
\subsubsection{Value-Based Methods}
The network is built using 3 fully connected layers. The input layer takes in 31 observations, the hidden layer has an input of 64 neurons and an output of 128 neurons, and the output layer has 9 nodes, to account for action space discretization. The input and hidden layers are activated using ReLU and the network is optimized using ADAM. 

\subsubsection{Policy-Based Methods}
The architecture for the Policy-Based methods is drawn directly from \cite{lillicrap_continuous_2015, fujimoto_addressing_2018}, with three fully connected layers; the input layer has 31 nodes, the hidden layer has an input of 400 nodes and an output of 300 nodes, and the output layer has 2 nodes to account for each wheel action. The input and hidden layers are activated using ReLU while the output layer is activated using a hyperbolic tangent. The network is optimized using ADAM.

\section{Experiments and Discussion} \label{Exp}

\subsection{Simulation}
All experiments were conducted on a computer with an Intel i7 processor and an NVIDIA RTX 3070 graphics card. We used the ARGoS multi-robot simulator \cite{pinciroli_argos_2012}, the Buzz swarm programming language \cite{pinciroli_buzz_2016}, and the PyTorch \cite{paszke_pytorch_2019} deep learning library for Python3 \cite{rossum_van_python_2009}. ARGoS sends observations from the environment via ZeroMQ to a Python server for learning via PyTorch. Actions are then chosen and sent back through ARGoS to Buzz for execution. Training occurred over 1,000 episodes, where each episode either ended by reaching the goal or by exceeding a time limit of 4,500 time-steps. The current weights of the model were saved every 10 episodes. Once trained, the best performing model was chosen and tested for 500 test cases.

\subsection{Hyper-Parameters}
We drew inspiration from \cite{mnih_human-level_2015} for our choice in hyperparameters. Specifically, a discount factor, $\gamma$, of 0.99997 and a learning rate $\eta$ of $10^{-4}$ were used. We used mini-batch learning with a memory of $10^6$ experiences and a batch size of 100 experiences, where an experience is comprised of the current state, action taken, reward gained, the next state, and a boolean terminal flag. We used an $\epsilon$-greedy learning approach with a linear decrement equal to $10^{-6}$ and a minimum $\epsilon$ value of 0.01. The target network was updated every 1,000 learning iterations. During training, learning occurred every time step.

\subsection{Scalability}\label{Scalability}

\begin{figure}
  \includegraphics[width=\linewidth]{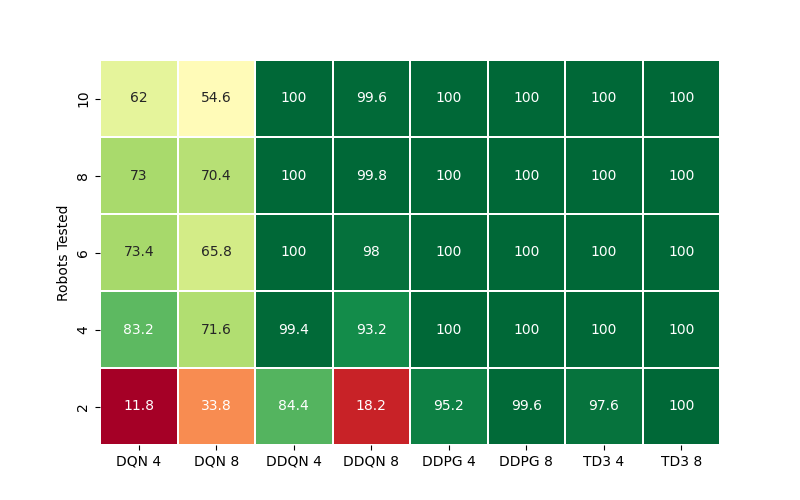}
  \caption{This heat map shows the results of our scalability study. The models that were trained, along with the designating number of robots that were trained, is shown on the x-axis. The y-axis is the number of robots tested with each trained model. The colors of the heat map are from red, representing worse scores, to yellow, representing mediocre scores, to green, representing good scores.}
  \label{fig:Scalability}
\end{figure}
Scalability is an important question in collective transport. Whether a method can scale or not is directly related to enabling the transport of larger, or smaller, objects than those that were used in training. To address this question, we trained two models, consisting of 4 robots and 8 robots, per method and tested them on 2, 4, 6, 8, and 10 robots. Each model was tested on 500 randomly generated configurations with the measure of success being if the object reaches the goal. The results are presented in Figure \ref{fig:Scalability}. 

Most notable is the poor performance of DQN relative to the rest. DQN shows dramatically poor scaling when scaling to 2 robots and shows a drop in performance when scaling up. We hypothesise this stems from DQN's overestimation of the Q-values, rendering a trained model unable to cope with a slightly changed configuration. 

DDQN, on the other hand, shows the ability to scale to higher numbers of robots, but also suffers from hampered performance scaling down, especially in the case of the model trained on 8 robots. This shows that value-based methods can scale with higher degrees of freedom over what they were trained with. However, they have a difficult time adjusting to under actuation. 

Both DDPG and TD3 show robust performance scaling. We do see a slight drop-off when scaling to 2 robots which could be associated with DDPG's tendency to overestimate state-action values. 

\subsection{Resilience}
\begin{figure*}
  \includegraphics[width=\linewidth]{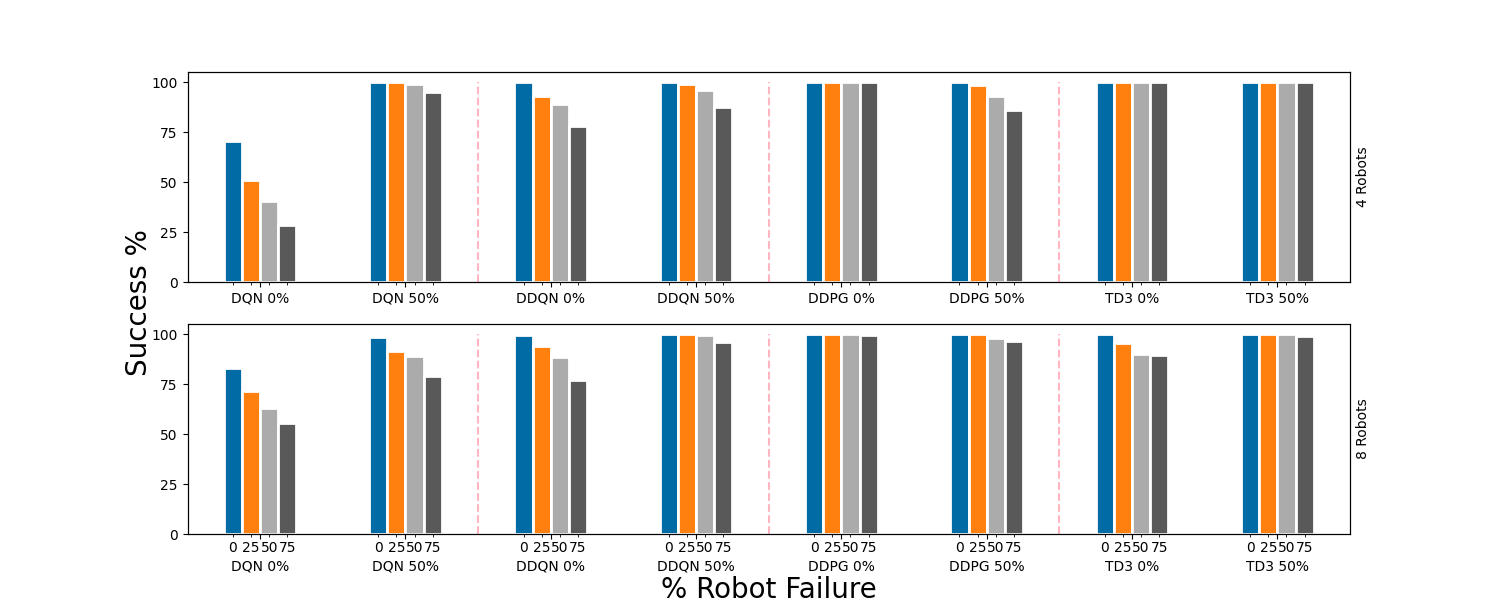}
  \caption{The results of the resilience study are presented here. The top and bottom plots correspond to models trained on 4 and 8 robots respectively. The x-axis shows blocks of 4 bars, where each block represents a model that was trained with either 0\% failure or 50\% failure. The 4 bars within each block represent the 4 tests performed on each model ranging from 0\% to 75\% failures. The y-axis corresponds to the success rate, where a success is defined by reaching the goal.}
  \label{fig:Resilience}
\end{figure*}

We study the effects of failure in the form of an attrition study. We analyze the effects of losing up to 25\%, 50\%, and 75\% of robots during testing. To prevent over-fitting on a specific number of failures, at the beginning of an episode we uniformly at random choose the number of failures that will occur, from a minimum of 0 up to the set theoretical maximum. So, for example, if we are training a model with 8 robots and 50\% failure, each episode can have at most 4 failures, but can have fewer. We define a failure as a complete loss of power, releasing the gripper so the robot is no longer attached to the object being transported and can no longer act upon it. Once a robot is determined to fail, we uniformly at random generate the time-step at which it will fail. Each method was trained with 4 and 8 robots and with 0\% and 50\% robot failures.

Figure \ref{fig:Resilience} shows our results, where the top and bottom plots correspond to 4 and 8 robots respectively. Each plot has 2 entries per method, 0\% and 50\% corresponding to the number of allowed failures during training. Within each bar set, we plot the results from testing on 0\%, 25\%, 50\%, and 75\% failures allowed. 

Policy-based methods, both trained with failures and without, perform better than value-based methods, especially as the failure percentage increases. We can also see that, within value-based methods, the models trained with failures present perform better than those without. The policy-based results are not as clear, but show resilience to failure in all configurations. 

Failure testing, while informative for understanding resilience, also allows us to study a system that is non-symmetrical. Our results indicate extending policy-based methods to non-symmetrical transport conditions may be effective, and show that value-based methods may suffer.

\subsection{Obstacle Avoidance}
Obstacle avoidance is an important part of any collective transport algorithm. Here we explore model adaptability via two obstacle types as shown in Figure \ref{fig:Obstacles}. 

\subsubsection{Cylinder Obstacles}
Cylinder obstacles are the same dimensions as the cylinder to transport and are generated uniformly at random within the area between the initial robot positions and the goal. We train in an environment with 0 and 2 obstacles and study the effective strategies in environments with both 2 and 4 obstacles. 

Table \ref{tab:Obstacles} reports our findings, where DQN-0 refers to DQN trained in an environment with 0 obstacles and DQN-2 refers to DQN trained in an environment with 2 obstacles. All methods show deterioration in performance in the 4 obstacle environment as compared to the 2 obstacle environment. This is due to an increased obstacle density within the environment. As we add more obstacles, the probability of introducing non-convex obstacles increases, as a result of proximity between two cylinder obstacles. Non-convex obstacles have proven significantly difficult to solve in the literature \cite{farivarnejad_multirobot_2022}. 

DQN suffers in both training configurations and in both obstacle environments, showing a drop in performance as we increase number of obstacles. We see relatively similar performance from DDQN, DDPG, and TD3. 

\begin{table*}[t]
  \begin{center}
    \caption{Success Rates with Cylinder Obstacles}
    \label{tab:Obstacles}
    \begin{tabular}{c|c|c|c|c|c|c|c|c}
      \toprule 
       Obstacles &DQN-0 &DQN-2 &DDQN-0 &DDQN-2 &DDPG-0 &DDPG-2 &TD3-0 &TD3-2\\
      \midrule 
       2  &64.4\% &73.2\% &93.6\% &90.4\% &90.2\% &91.4\% &92.4\% &91\%\\
       4  &48.8\% &56.2\% &79.8\% &81.2\% &83.2\% &86.4\% &84\% &82\%\\
      \bottomrule 
    \end{tabular}
  \end{center}
\end{table*}

\begin{figure}
  \centering
  \includegraphics[width=90mm]{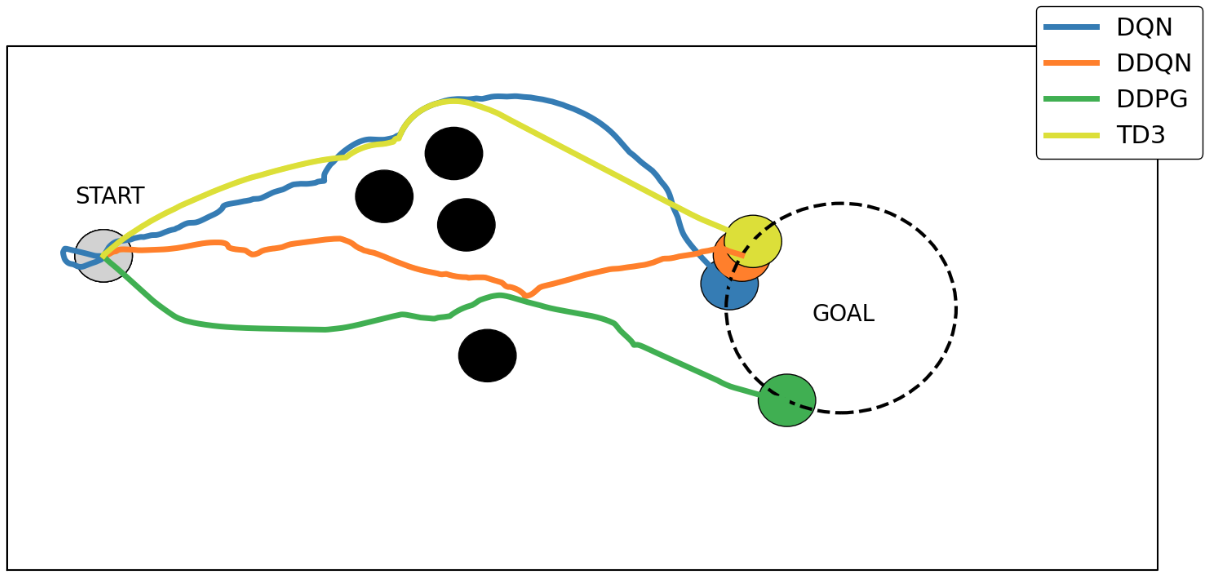}
  \includegraphics[width=90mm]{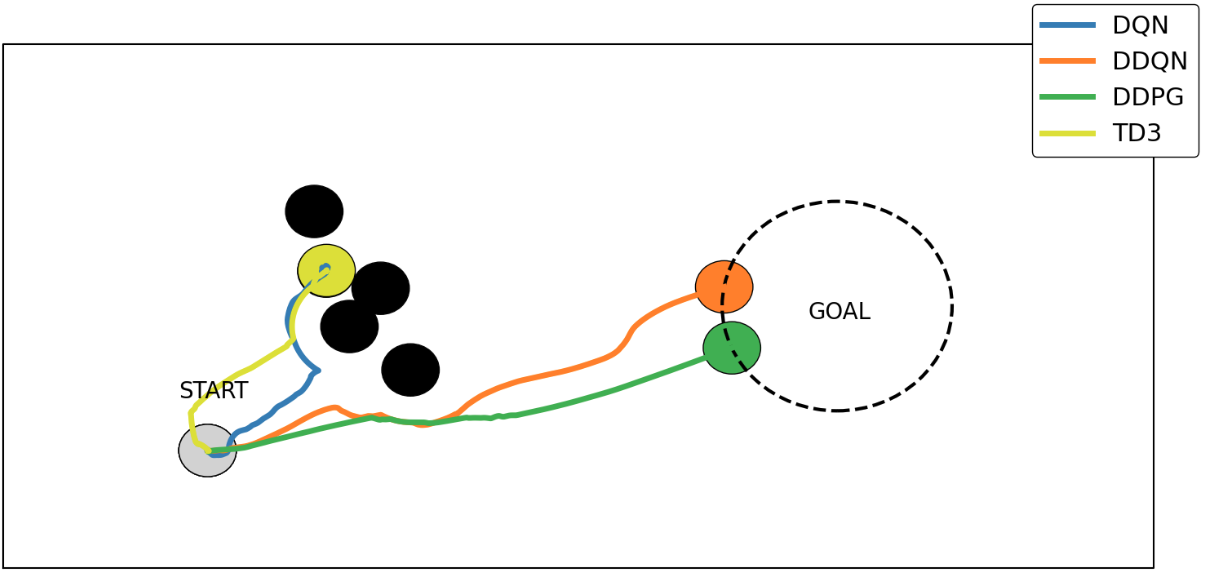}
  \includegraphics[width=90mm]{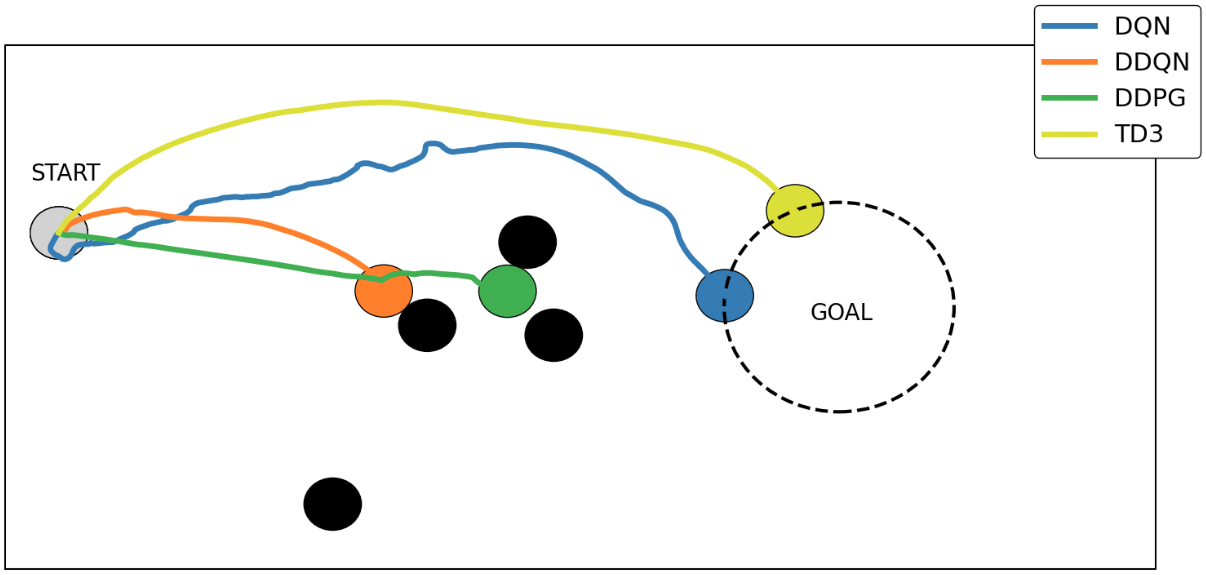}
  \caption{The trajectories of the best neural networks trained with 2 obstacles and tested with 4 obstacles. The top plot shows successful trials for each learning algorithm considered. DQN and TD3 selected strategies that move above the obstacles, while DDQN and DDPG produced strategies that move below the obstacles. The middle plot is an example of a failed attempt at overcoming the obstacles with DQN and TD3. The bottom plot shows failures for DDQN and DDPG.}
  \label{fig:4_Obstacles}
\end{figure}

Figure \ref{fig:4_Obstacles} shows trajectories from the 4-obstacle environment. The top plot shows the difference in behaviors learned. DQN and TD3 learned to maneuver towards the top of the environment as they made their way towards the goal, whereas DDQN and DDPG learned to maneuver towards the bottom of the environment. This differing behavior only emerged in the environment with obstacles. In all other environments we see similar strategies between methods. We also notice the smoothness in trajectories where DQN is the least smooth followed by DDQN, DDPG, and TD3 which suggests an increased level of implicit coordination amongst the robots. Less smooth trajectories point to disagreement in obstacle trajectory between robots leading robots to apply forces in different directions. In contrast, smoother trajectories points to all the robots applying similar forces, thus having a higher level of implicit coordination.

The middle and bottom charts are examples of a non-convex obstacle developed by the proximity of two cylinder obstacles being too small for the robots to pass through, yet large enough that the robots are not able to pass over it. All methods struggle with this type of obstacle.

\subsubsection{Gate Obstacle}
The gate obstacle represents a more difficult obstacle as the robots must find an opening in the wall structure to get to the goal. Robots have no way to explicitly communicate, so a single robot sensing the opening may not result in a successful trial.

The gate obstacle adds a layer of complexity to the task in that the optimal policy is no longer strictly to move towards the goal. The models must now learn to wall-follow to find the opening and then find the goal. DQN and DDQN were able to reach the goal 72.4\% and 74.4\% respectively, whereas DDPG and TD3 performed better, reaching the goal 77\% and 76.4\% respectively.
\begin{figure}
  \centering
  \includegraphics[width=90mm]{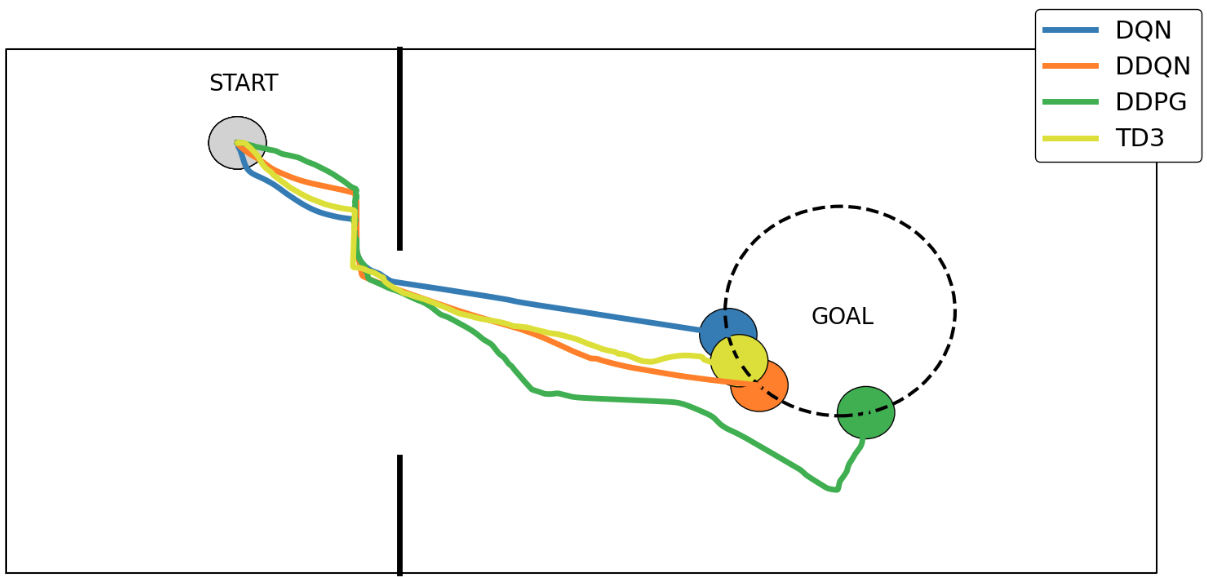}
  \includegraphics[width=90mm]{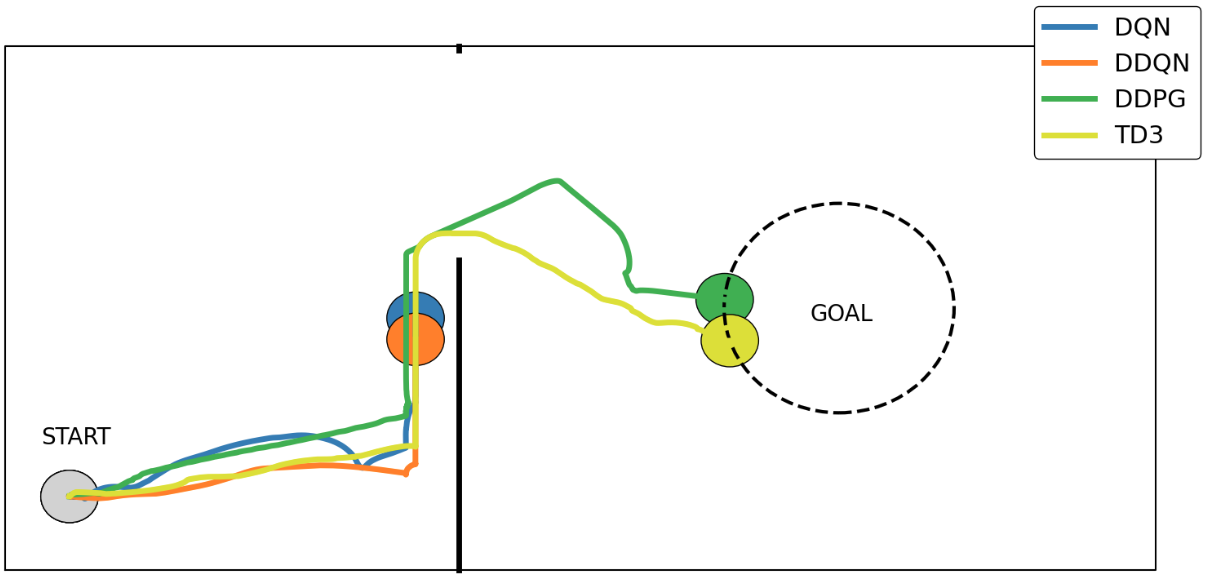}
  \caption{The trajectories of the best neural networks trained on the gate obstacle. The top plot exemplifies a wall following behavior to overcome the obstacle and reach the goal. The bottom plot shows the disparity between Value-Based methods and Policy-Based methods. Value-Based methods are unable to overcome the obstacle if the opening is beyond the midpoint of the goal opposite from the starting position.}
  \label{fig:Gate}
\end{figure}

Figure \ref{fig:Gate} shows the behaviors learned in the gate obstacle environment. In the top chart we see a noticeable wall following behavior from all the methods. DQN takes the most direct path to the goal after completing the obstacle. This is due to a naive policy of simply moving towards the goal at all times. DDQN employs a similar policy. While this policy may work in some configurations, it fails when the opening of the gate is on the opposite of the mid point of the goal from the robots. This is evident in the bottom plot of Figure \ref{fig:Gate}. TD3 and DDPG are able to continue in the up direction until they find the opening whereas DQN and DDQN stop at the mid point of the goal.


\section{Conclusion and future work}
We presented a study of four well-known RL algorithms applied to aggregates of minimalistic robots. We show that implicit communication is a viable method to solve non-stationarity in multi-agent reinforcement learning. We demonstrated this insight in a collective transport scenario, in which the robots are physically connected to the object to transport. The control strategies that resulted from the training methods we compared generally showed the ability to scale, be resilient to failures, and coordinate around obstacles, although with different levels of success.

The idea of implicit coordination is compelling and future work will aim to develop it further. While in this paper we considered objects whose shape is radially symmetric, our method could be expanded to support arbitrarily shaped objects, and even deformable ones. This could be obtained by explicitly learning the behavior of the object to transport in a dedicated module. Eventually, this methodology could be applied to other types of minimalistic aggregates, such as modular robotic platforms.


\bibliographystyle{IEEEtran}
\bibliography{IROS2022}

\end{document}